\documentclass[11pt,a4paper]{article}
\usepackage[hyperref]{acl2021}
\usepackage{times}
\usepackage{latexsym}

\usepackage{microtype}
\aclfinalcopy

\usepackage{hyperref}
\usepackage{graphicx}
\usepackage{float}
\usepackage[graphicx]{realboxes}

\usepackage{geometry}
\usepackage{enumitem}

\title{\Large Model Bias in NLP\\  
 Application to Hate Speech Classification using transfer learning techniques}

\author{Aygul Zagidullina \\
  ETH Zürich \\\And
  Georgios Patoulidis \\
  ETH Zürich \\\And
  Jonas Bokstaller\\
  ETH Zürich \\}

\begin{document}
\maketitle
\begin{abstract}
In this paper, a BERT based neural network model \citep{devlin2019bert} is applied to the JIGSAW data set \citep{real_data} in order to create a model identifying hateful and toxic comments (strictly seperated from offensive language) in online social platforms (English language), in this case Twitter. Three other neural network architectures and a GPT-2 \cite{gpt2} model are also applied on the provided data set in order to compare these different models. The trained BERT model is then applied on two different data sets to evaluate its generalisation power, namely on another Twitter data set \citep{hatespeech_twitter} \citep{hateoffensive} and the data set HASOC 2019 \citep{hasoc2019} \citep{10.1145/3368567.3368584} which includes Twitter and also Facebook comments; we focus on the English HASOC 2019 data. In addition, it can be shown that by fine-tuning the trained BERT model on these two data sets by applying different transfer learning scenarios via retraining partial or all layers the predictive scores improve compared to simply applying the model pre-trained on the JIGSAW data set. With our results, we get precisions from 64\% to around 90\% while still achieving acceptable recall values of at least lower 60s\%, proving that BERT is suitable for real use cases in social platforms.
\end{abstract}

\section{Introduction}

Toxic comments severely affect public conversations on the internet. Recently, there have been several research initiatives that aim at detecting such comments using algorithms, to name a few, the social science research projects \href{https://ethz.ch/services/en/news-and-events/internal-news/archive/2021/03/an-algorithm-that-detects-hate-speech.html}{“Stop Hate Speech”} \cite{stophatespeech_ethz} at the ETH Zürich and \href{https://dh.berkeley.edu/projects/online-hate-index-ohi-research-project?utm_source=BCNM+Subscribers&utm_campaign=d5d78bba5e-natasha-schull-oct12_COPY_01&utm_medium=email&utm_term=0_eb59bfff9e-d5d78bba5e-281420185}{“Online Hate Index”} \cite{stophatespeech_cal} at the University of California, Berkeley. Moreover, initiatives to protect voices in public discussions were also undertaken in the technology industry, for example, by the Conversation AI team (founded by Google Jigsaw).  
\newline
\newline
A main area of focus is machine learning models that can identify toxicity in online conversations, where toxicity is defined as anything rude, disrespectful, or otherwise likely to make someone leave a discussion. However, there should be made a differentiation between hate speech and offensive language. It is often the case that certain comments on the internet contain an offensive lexicon (e.g. cursing, disrespectful communication), however, they are not a hate speech \textit{per se} (i.e., speech which is used to attack people based on characteristics like race, ethnicity, gender, and sexual orientation, or threats of violence towards others \footnote{The definition is according to the Facebook's policy, \href{https://m.facebook.com/communitystandards/hate_speech/}{Facebook.Community Standards. Hate Speech}}). Thus, given the data from available resources it is challenging to disentangle the actual hate speech from the offensive language usage. Furthermore, there are comments that do not explicitly use hate or offensive language but still attack certain identities. The machine learning models trained on this data are prone to a classification bias. For example, the model can predict a high likelihood of hate speech for comments which actually contain abusive language. As a consequence, certain individuals might potentially be censored from online conversations based on the binary (hate vs. non-hate) detection algorithms.
\newline
\newline
In this project, we will develop a model that separates hate speech from the offensive and non-hate/non-offensive language, and hence, minimizes the bias; and transfer the learned model to unseen Twitter/Facebook data. Thus, we aim at building a generalized model that operates fairly across a diverse range of conversations with correct categorization of hate speech.

\section{Background}

The issue of hateful and toxic comments on social platforms has lead to an ongoing controversy regarding censorship and enforcing platform rules in such communities. Many attempts were done to automatically detect such comments to support humans in this task, a necessary step as the growing number of such comments make a manual checking approach impossible. Nevertheless, algorithms are not perfect, can be biased and thus can occasionally lead to the deletion of comments which clearly don't contain any hate speech. For example, criticizing a specific government or a specific policy is by no means hate speech; censoring such comments may contribute to political tensions. It is in the best interest of social media platforms which are run by profit-oriented companies, largely from the USA, to minimize such censoring in order to serve shareholders by generating sustainable profits. The reason is clear: The political mainstream and sentiment can change fast and such companies may be targeted one day for censoring specific opinions. In addition, decentralised solutions might make such companies obsolete, one example of 'creative destruction' as termed by Joseph Alois Schumpeter. There are multiple projects and start-ups working on such solutions, for example on the Ethereum network \citep{8645058} \citep{wblockchain}. Lastly, the growing political division might lead to different social platforms with their own biases by excluding specific opinions where users get a pretended view on societal reality: One social platform might tend to censor opinions in favor of a certain government and another one against it.
\newline
\newline
There has been done some research in this field and social media companies are exploring their own algorithms. Nevertheless, there is an issue with precision of predicted results, as can be clearly seen by the growing numbers of criticism against excluding certain opinions from social platforms but also by reviewing literature. For example, \citep{10.1145/3368567.3368584} report no clear precision at all but rather report a weighted F1 score in the range of 70\% - 80\%, suggesting that precision is worse off than recall as that property is reported by \citep{hateoffensive}: their best performing model achieves an overall precision of 91\% and a recall of 90\%, but a closer look at the sensitive part reveals that precision and recall for the hate speech class of comments is only 44\% and 61\%. This is not suitable for a productive and live version for a social media platform, as precision is the measure that counts most in this sensitive topic, which should be chosen over recall in the precision-recall trade-off. Thus, the objective of this work is to evaluate if better scores can be achieved.

\section{Data and preprocessing}

We create a pipeline that automatically downloads, extracts and preprocesses the dataset from Kaggle \cite{real_data} and creates a training/testing data set for different models (we use a train/test split of 0.1). 

\subsection{Real training data}
\begin{figure}
\centering
\includegraphics[scale=0.5]{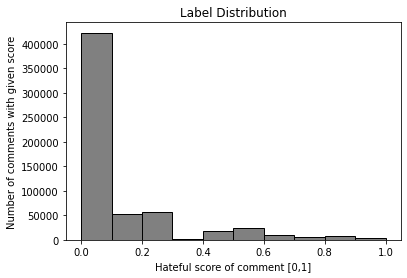}
\caption{Distribution of scores (X) assigned to the comments. The higher the score the more hateful content the comment contains.} \label{fig3}
\end{figure}

We use the publicly available \href{https://www.kaggle.com/c/jigsaw-unintended-bias-in-toxicity-classification/data}{Jigsaw/Conversation AI data set} \cite{real_data} from the Kaggle competition. The data contains about 2 mio. public comments with an extended annotation by human raters for various toxic conversational attributes \footnote{At the end of 2017 the Civil Comments platform shut down and chose make their 2 mio. public comments from their platform available in a lasting open archive so that researchers could understand and improve civility in online conversations for years to come. Jigsaw sponsored this effort and extended annotation of this data by human raters for various toxic conversational attributes.}. Each text comment in the train data has a toxicity label (very toxic, toxic, hard to say and not toxic). This attribute is a fractional value which represents the fraction of human raters who assigned the attribute to the given comment. For evaluation purposes, test data examples with target $> =$ 0.5 are considered to be in the toxic class (i.e., very toxic, toxic) vs. non-toxic (i.e., hard to say and non-toxic).
Furthermore, this training data is pre-processed in order to get structured text input that can be worked with; the free text fields are cleaned up by removing special characters and stop words, after that tokenization and lemmatization are applied.
\newline
\newline
From this data sets we extract the text and hate speech score which we use as a target variable. In total we have 1'804'874 comments with according scores (see Figure \ref{fig3}).

\subsection{Real transfer data}
\label{marker}

We test our trained model on a data set containing carefully labeled Twitter tweets with regards to hate or offensive language and neither of both \cite{hatespeech_twitter} \cite{hateoffensive}. It is important to achieve a high precision as false predictions of hate speech leading to censorship in social media contributes to the controversy of inflicting with first amendment rights. In addition, we compare our trained model with another English labelled data set - HASOC 2019 \cite{hasoc2019} \cite{10.1145/3368567.3368584} - which also includes the differentiation in labels hate speech, offensive and profane language and neither using Twitter and also Facebook comments. 
\newline
\newline
\cite{hatespeech_twitter} data has about 27'000 comments with the ‘hate speech’ (class 0), ‘offensive language’ (class 1) and ‘neither’ (class 2) labeling. We remove rows containing class 1 labeling. The resulting data has ca. 5'600 rows.  
\newline
\newline
The HASOC 2019 data set has about 6'000 comments with the with labels hate, offensive or profane language and neither of those. We use the label ‘HATE’ as hate, ‘NONE’ as non-hate based on the labeling of task 2, and exclude rows containing ‘OFFN’ (offensive) and ‘PRFN’ (profane) labels. The resulting data is ca. 4'700 rows. \\
\newline
Such filtering of the transfer data sets has been performed in order to ensure the alignment with the training data. Since the models are trained only on hate and non-hate labels, testing the model on more labels is not fair and yields worse performance (the additional results confirming this observation are available upon request).

\subsection{Preprocessing}

\begin{figure}
\centering
\includegraphics[scale=0.5]{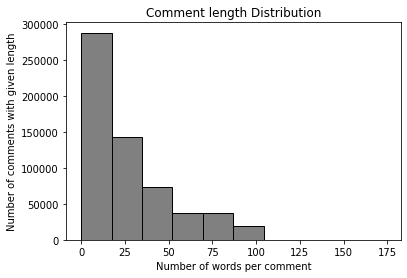}
\caption{Distribution of the lengths of all comments after preprocessing.}\label{fig2}
\end{figure}

To begin with, the stop words have to be removed. We use the NLTK library's \citep{NLTK} standard stop-word list and also include our own regular expression matching to remove punctuation, non-alphabetic characters, URLs, HTML characters, twitter tags, non-ASCII characters and text between square brackets. Also, all upper characters are replaced by lower cases. After these steps, we make use of NLTK's lemmatiser to finalise the preprocessing for a given data set containing free text fields. Furthermore, we analyze the typical number of words in each sentence after prepossessing (see Figure \ref{fig2}). Based on this analysis we use a maximum length of 128 words, and when applying tokenization to the texts we pad tokenized outputs to 128 words per comment.

\subsection{Preprocessing: Deep Learning Models}

For the Deep Learning architectures created based on GRU and LSTM, we use as embedding the pretrained GloVE \cite{pennington2014glove} model with 840B tokens and 300d vectors to have as many matching words as possible. Based on this model we build the embedding matrix based on the tokenizer's dictionary and use this matrix for the first layer in all our models as an embedding layer. Thus, we have a matrix (i, 128, 300) as input, where i represents the batch size, 128 is the number of words per comment and 300 is the embedding vector for each word.

\subsection{Preprocessing: Transformers}

For transformer models we use the same cleaning procedure as described above. However, we use special tokenizers of these transformer models. Compared to word embeddings, BERT and GPT-2 \cite{gpt2} models' tokenizer includes an attention mask which is used for the attention mechanism that is able to learn contextual relations between words in text. 
\newline
\newline
We used the following tokenizers from \href{https://huggingface.co/transformers/pretrained_models.html}{HuggingFace's Transformers} module:
\begin{itemize}[leftmargin=*]
  \item BERT tokenizer: \textbf{BertTokenizerFast} for model specification bert base uncased;
  \item GPT-2 tokenizer: \textbf{GPT2TokenizerFast} for model specification gpt2.
\end{itemize}

\subsection{Training Data Size}

For the model training we use only the first 600 000 observations of our entire JIGSAW data set due to the computational burden. However, this does not affect the accuracy metrics (defined later in Section 5) of our model. For completeness, we have tested the performance of the transformer model (BERT) on different subsets of data, i.e. 600 000, 1200 000 or all observations \footnote{Training of the model given all observations is based only on 2 epochs instead of 3 compared to the other 2 subsets.}. The differences are minor with the marginal improvement as data size grows (which is expected). This is explained by the fact that the class imbalances are stable across subsets of observations (see Section below and Table \ref{table1}).  

\begin{table}[ht]
\caption{Performance metrics for BERT (validation set)} 
\centering 
\Rotatebox{270}{%
\begin{tabular}{l c c c c} 
\hline\hline 
Model & accuracy & precision & recall & AUC \\ [0.5ex] 
\hline 
BERT 600 k obs & 0.95 & 0.70 & 0.64 & 0.81 \\ 
Bert 1200 k obs  & 0.95 & 0.72 & 0.63 & 0.81 \\
BERT all obs & 0.95 & 0.73 & 0.61 & 0.80 \\ [1ex] 
\hline 
\end{tabular}
}%
\label{table1} 
\end{table}

\subsection{Data Properties: Class Imbalance}

Class imbalance for the JIGSAW data set is about 92\% for non-toxic and 8\% for toxic (hateful) comments irregardless of the number of observations used, i.e. 600 000 or 1200 000 or the entire data set. 
\newline
\newline
Class imbalance for the \cite{hatespeech_twitter} data set is 74.4\% non-hate and 25.6\% hate. In turn, class imbalance for the HASOC 2019 data is 75.6\% non-hate and 24.4\% hate. These numbers are with regards to our label extraction (only hate as such and non-hate and non-offensive as the opposite).

\section{Methods}

Our main task is to perform transfer learning (without retraining of final layers) on two different data sets. Our model is trained on the JIGSAW data set \citep{real_data} and applied further to the new unseen data in order to test the generalization performance. The whole code is written in Python and given the complexity of the task, the code was run via a Google Colab notebook using \href{https://cloud.google.com/tpu/docs/tpus}{TPUs}.

\subsection{Neural Network Architectures}

\subsubsection{Deep Learning Models}

\begin{figure}
\centering
\includegraphics[scale=0.25]{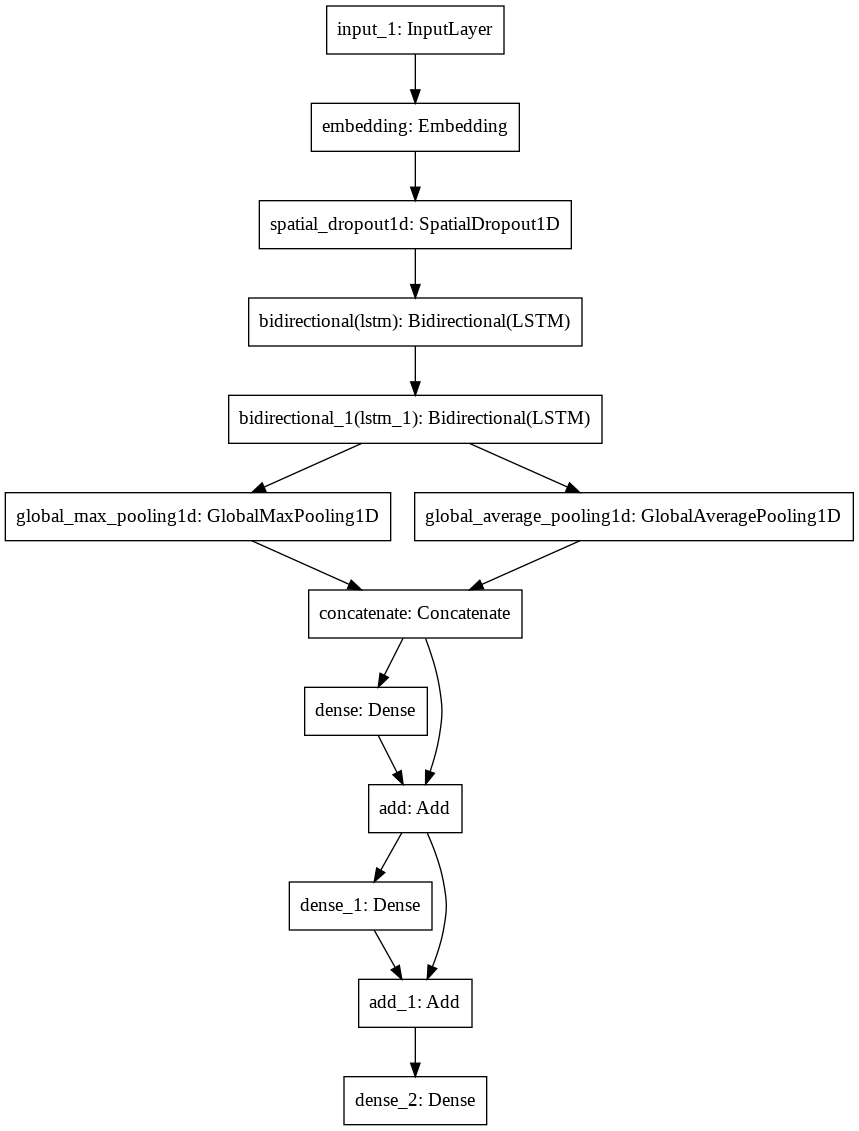}
\caption{Model structure of the sophisticated model that benefits from the nature of hateful speech comments containing trigger words.}\label{fig1}
\end{figure}

The Deep Learning models comprise two simple RNN models, one with LSTM cells and one with GRU cells, and a more sophisticated model \cite{advanced_model}. The two simple models use a standard model structure and don't use properties of the hate speech nature as for example triggering words implying indeed a toxic comment.
\newline
\newline
The sophisticated model has a more complex structure (see Figure \ref{fig1}) but also uses LSTM layers (bidirectional LSTM layers \cite{lstm}) at its heart. We start by applying the typical embedding layer from the GloVe data set \cite{pennington2014glove}. To avoid over-fitting, we add a dropout layer, followed by two bidirectional LSTM layers \cite{lstm}. After that the two pooling layers are used in parallel by utilizing the Keras functional API which allow more complex structures and not only sequential ones.
The left branch pools the maximum value from the previous LSTM layer and the right one the average value. This has the benefit that the model can learn the notion of trigger words (e.g. "silly", "stupid", etc.) which have a strong effect on the final prediction score. After this the two layers are concatenated followed by tree dense layers, one of them outputting the final prediction value.
The last dense layer uses as activation function the sigmoid to obtain a score between 0-1 that resembles how likely the comment contains hateful speech.
\newline
\newline
All three models use a batch size of 512 and are trained over three epochs. For gradient descent, the Adam (Adaptive Moment Estimation) \cite{kingma2017adam} optimizer is used with the binary cross-entropy as error function.

\subsubsection{Transformers}

Transformer models are currently very popular methods used to solve various NLP tasks such as question answering, language understanding or summarization, but they also have been successfully used in text classification tasks (see e.g. \cite{bert_appl}). \\
\newline
BERT (Bidirectional Encoder Representations from Transformers) is a language transformation model introduced by Google \cite{devlin2019bert}. BERT is “deeply bidirectional”, which means it learns the deep representation of texts by considering both, left and right contexts. It is a method used for training general purpose language models on very large corpuses of data and then using that model for the NLP tasks. So there are two steps involved in using BERT: pre-training and fine-tuning. During the pre-training phase, the BERT model is trained on a large corpus of data. Then, the model is initialized with the pre-trained parameters and fine-tuned for specific NLP task. Fine-tuning of the BERT model is much less expensive on the computational resources. \\
\newline
BERT uses the same architecture in different tasks and is built using Transformers \cite{transformers}. The basic model is BERT base uncased, it consists of 12 transformer blocks, hidden units size of 768 and 12 self-attention heads, and trained on lower-cased English text. The total number of parameters comprises 110 million. \\
\newline
We use the basic BERT model as stated above with the pre-trained parameters and fine-tune it to our specific JIGSAW data set by training this model for 3 epochs. To define a neural network, we use the "AutoModelforSequenceClassification" model from HiggingFace's Transformers module where we set the \emph{model\_checkpoint} parameter to \textbf{bert-base-uncased}. \\
\newline
Given that we "re-train" 110 million parameters, this task is computationally cumbersome. Hence, it was decided to use the \href{https://huggingface.co/docs/accelerate/}{HuggingFace's Accelerate API} that allows running the model in distributed setting, on 8 TPUs \footnote{ The Tensor Processing Unit (TPU) is a custom ASIC chip designed from the ground up by Google for machine learning workloads.}. For the data set size of 600 000 observations (which is one third of the entire data set), the training of the model takes approximately 4.5 hours. 
\newline
Another transformer model which we use is OpenAI GPT-2 model that was proposed in \cite{gpt2}. It’s a causal model pre-trained using language modeling objective (predicting the next word given a sentence) on a very large corpus of text data. This model comprises 12 transformer blocks, hidden units size of 768 and 12 self-attention heads which yields 117 million parameters.\\
\newline
In order to define a neural network, we use the "GPT2ForSequenceClassification" model from HiggingFace's Transformers module where we set the \emph{model\_checkpoint} parameter to \textbf{gpt2}. We train this model for 3 epochs on 8 TPUs using Accelerate API as well.

\subsection{Tuning Parameters and optimization}

We define a DataLoader function that results in a 8-batched size output for training and testing data which in turn is an input for the neural network. The above described neural network architectures have the following common user-defined parameters:
\begin{itemize}[leftmargin=*]
 \item \emph{learning\_rate} = 0.00001 \vspace*{-0.25cm}
 \item \emph{num\_epochs} = 3 \vspace*{-0.25cm}
 \item \emph{train\_batch\_size} = 8 (actual batch size is 64 because of the data input function mentioned above) \vspace*{-0.25cm}
 \item \emph{eval\_batch\_size} = 8 (actual batch size is 64 because of the data input function mentioned above) \vspace*{-0.25cm}
 \item \emph{seed} = 42 (for reproducibility)
\end{itemize}
The tuning parameters specific to the Neural Network architectures are left out for the sake of concision, however, they are available in the respective Colab python notebooks and upon request.
\newline
\newline
To minimize the error rate of the model in the prediction, we use the Adam \cite{kingma2017adam} optimization technique. It computes the learning estimation for each parameter. Adam yields faster convergence and training speed, also it alleviates  high variance of the parameters.

\section{Performance Metrics}

To evaluate the models, we decided to use the standard metrics used in classification, e.g., accuracy, precision and recall. Such metrics are easy and straightforward to obtain for a binary classification problems and can be computed as:

\begin{itemize}
    \item Accuracy = $\frac{TP + TN}{TP + FP + FN + FP}$
    \item Precision = $\frac{TP}{TP + FP}$
    \item Recall = $\frac{TP}{TP + FN}$
\end{itemize}
where
\begin{itemize}
    \item TP—True Positive: examples are predicted to be positive and are positive
    \item TN—True Negative: examples are predicted to be negative and are negative
    \item FP—False Positive: examples are predicted to be positive but are negative
    \item FN—False Negative: examples are predicted to be negative but are positive.
\end{itemize}

To compare the models, we also use the Area Under the Receiver Operating Characteristic Curve (AUROC, short AUC) score to evaluate the models.
Although the AUROC score is not an ideal metric to compare the models trained on highly imbalanced data, we use it
to be aligned with the relevant literature as the most studies use the AUROC score.

\section{Experimental Findings}

\subsection{Training Data Results}

For a fair model comparison, we use the same number of observations and tuning parameters for all 5 models under consideration. The predictive metrics are provided in Table \ref{table2}.

\begin{table}[hbt!]
\caption{Performance metrics comparison of models (validation set)} 
\centering 
\Rotatebox{270}{%
\begin{tabular}{l c c c c} 
\hline\hline 
Model & accuracy & precision & recall & AUC \\ [0.5ex] 
\hline 
Basic LSTM Net & 0.94 & 0.74 & 0.48 & 0.89 \\ 
Basic GRU Net & 0.94 & 0.70 & 0.48 & 0.92 \\
LSTM Pooling Net & 0.94 & 0.65 & 0.43 & 0.92 \\ 
BERT Net & 0.95 & 0.70 & 0.64 & 0.81 \\
GPT-2 Net & 0.94 & 0.68 & 0.57 & 0.77 \\ [1ex] 
\hline 
\end{tabular}
}%
\label{table2} 
\end{table}

\subsection{Test Data Results}

Given that on our training JIGSAW data set, the BERT model overall performs best, we run this fine-tuned model on the test data sets to evaluate the generalization performance. For comparison, we also run GPT-2 model on the test data sets. The results are provided in Table \ref{table3}. 

\begin{table}[hbt!]
\caption{Performance metrics comparison of BERT and GPT-2} 
\centering 
\Rotatebox{270}{%
\begin{tabular}{l c c c c} 
\hline\hline 
Model BERT & accuracy & precision & recall & AUC \\ [0.5ex] 
\hline 
Davidson data & 0.83 & 0.63 & 0.78 & 0.81 \\ 
HASOC data & 0.59 & 0.23 & 0.28 & 0.49 \\ [1ex] 
\hline\hline 
Model GPT-2 & accuracy & precision & recall & AUC \\ [0.5ex] 
\hline 
Davidson data & 0.79 & 0.59 & 0.63 & 0.74 \\ 
HASOC data & 0.60 & 0.22 & 0.26 & 0.48 \\ [1ex] 
\hline 
\end{tabular}
}%
\label{table3} 
\end{table}

\subsection{Transfer Learning}
To evaluate the generalisation power of our BERT model by putting it into context, three transfer learning approaches are considered. The first one is applied by re-training the parameters in all layers with regards to the Davidson and HASOC data, respectively. The second approach is based on a freezing of the embedding parameters while re-training the parameters in remaining layers. The third approach uses a freezing of the embedding parameters and parameters in 50\% of the layers for the BERT encoder part, while re-training the remaining layers' parameters. For this, a training/test split of 0.1 is used with 3 epochs again. The results are summarized in Table \ref{table4}.

\begin{table}[hbt!]
\caption{Performance metrics of the transfer learning for BERT} 
\centering 
\Rotatebox{270}{%
\begin{tabular}{l c c c c} 
\hline\hline 
Model BERT (frozen embeddings) & accuracy & precision & recall & AUC \\ [0.5ex] 
\hline 
Davidson data & 0.93 & 0.85 & 0.87 & 0.91 \\ 
HASOC data & 0.78 & 0.50 & 0.09 & 0.53 \\ [1ex] 
\hline\hline 
Model BERT (frozen embeddings  & accuracy & precision & recall & AUC \\ [0.5ex] 
 and 50\% of encoder layers) &  &  &  &  \\ [0.5ex] 
\hline 
Davidson data & 0.93 & 0.85 & 0.84 & 0.90 \\ 
HASOC data & 0.77 & 0.83 & 0.04 & 0.52 \\ [1ex] 
\hline\hline 
Model BERT (all layers) & accuracy & precision & recall & AUC \\ [0.5ex] 
\hline 
Davidson data & 0.94 & 0.92 & 0.85 & 0.91 \\ 
HASOC data & 0.74 & 0.60 & 0.07 & 0.53 \\ [1ex] 
\hline 
\end{tabular}
}%
\label{table4} 
\end{table}

\section{Analysis}
\subsection{Training}
The three basic neural networks (first three rows of Table \ref{table2}) show strong results on the 10\% hold-out validation set. Accuracy is consistently at 94\% with acceptable precisions, where the Basic LSTM Net model achieved the highest precision score of all five models at 74\%. The AUC scores of these three models also show strong results, valued around 90\%. Nevertheless, these models have recalls below 50\% in the 40s\% range which is not usable for a productive setting. For this given reason we didn't follow these models any further as the large models, namely BERT and GPT-2 show better results: being competitive with achieved accuracies (BERT also achieved the highest score with 95\%), the precisions are acceptable with BERT at 70\% and GPT-2 at 68\% and also the reported recall metrics, especially for BERT with 64\% compared to 57\% for GPT-2. The AUC is at around 80\% what seems to be bad compared to the three models - these results show the importance of reviewing many metrics as the overall picture is strong for BERT. That is why this models has our largest focus in this work. The accuracy paradox clearly applies here; a fact whose importance cannot be stressed often enough.

\subsection{Application to Davidson and HASOC 2019 data}
As justified in the subsection before, the three first neural networks of Table \ref{table2} are not followed any further. Even though GPT-2 performed worse than BERT (especially on recall, which is the tie-breaker so to speak), we include this model evaluated on the two test sets whose results can be seen in Table \ref{table3}. Both models show acceptable results for the Davidson data, with BERT being clearly superior over GPT-2 across all scores, especially for recall again. Interestingly, besides the expected decrease of predictive scores, AUC stayed the same and recall even improved from 64\% on the JIGSAW validation set to 78\% on the Davidson data. The improvement of recall can also be observed for the GPT-2 model which rises from 57\% to 62\%.
\newline
\newline
Applied on the HASOC 2019 data, both models perform bad with the impression of being completely random: AUC scores are just below 50\% and the other metrics are almost equal for both models, with recall and precision in the 20s\% range and accuracies of roughly 60\%.

\subsection{Transfer learning}
By focusing only on BERT now for the reasons already provided (please refer to Table \ref{table4} for the results), re-training this model while keeping the embedding parameters frozen and also 50\% of the encoder layers clearly improves all metrics by substantial margins for the Davidson data: accuracy improves from 83\% to 93\%, precision from 63\% to 85\%, recall from 78\% to 84\% and AUC from 81\% to 90\% (by comparing Table \ref{table3} and \ref{table4}) - using the same approach but only freezing the embedding layers yields in almost the same results. In addition, by re-training all layers the model metrics improve in a comparable way, whereas the largest difference can be seen for precision which even increased to 92\%. This is a clear finding of the strong generalisation power of the applied BERT model, as the frozen embeddings still allow great results on new unseen data.
\newline
\newline
Applying the same approaches to the HASOC 2019 data, one might be mislead to conclude an improvement of predictive power, but a closer look leads to the conclusion that this is not the case: recall drops to below 10\% for all three scenarios (even to 4\% for the frozen embeddings and half of the encoder layers). Even though accuracy increases by 19\% in the frozen embeddings case, by 18\% in the frozen embeddings and half of encoder layers and by 15\% for re-training on all layers and even that precision also increases to more than 20-30\%, these metrics are practically useless as the reported recalls are not acceptable under any circumstances. AUC still remains almost at random with with lower 50s\% for all transfer learning scenarios.

\section{Discussion}

Our results can probably be enhanced by tackling the class imbalance of the training data which we didn't do as the absolute amount of data points seemed convincing enough; this can be done by weighting class labels with respect to their relative share of the total data. Another bottleneck is the computational power available as training with the whole Jigsaw data set \citep{real_data} and using better computational clusters is an approach to follow for future evaluations, especially for training large models on multi-millions of comments. Our reported accuracies can compete with the Kaggle competition, but due to the given circumstances of such a project we did not use a possible computation infrastructure to the full extent. Another take-away is the handling of hashtag texts from Twitter comments: the missing empty spaces between words make it difficult for our models to differentiate properly, thus a more sophisticated pre-processing pipeline taking this issue into account is recommended for future research. Nevertheless, we are able to beat the precision and recall metrics for hate speech from the Davidson data in the corresponding work \citep{hateoffensive}: our BERT model consistently beats their reported precision of 44\% and recall of 61\%, in the plain evaluation of the JIGSAW data trained model and also in the transfer learning scenarios. From this, we conclude that carefully training hate speech as such with high quality labels to avoid confusion with offensive language can severely improve model performance.

Regarding transfer learning, we can show that retraining even with frozen embeddings and frozen 50\% encoder layers lead to better results than applying the plain pre-trained model. Even though this plain application of the model results in productive results, fast and flexible re-training of pre-trained large models is recommendable for productive solutions. It might also be worth to differentiate the setting: Are the free texts on hand from social media platforms like Twitter, Facebook or reddit? Are they from student platforms or specific forums? It is clear that hate speech is more primitive on Twitter compared to scientific forums for example; tackling this circumstance might improve the detection of more sophisticated hate speech attempts on large scale social platforms like Twitter.
\newline\newline
It should also be mentioned that the bad performance of our models applied on HASOC 2019 \citep{hasoc2019} can be explained by bad labelling; after manual scanning through our pre-processed data set and filtered for hateful comments, we can confirm that many comments are no hate speech at all but more sceptical and/ or populist political comments. Here are a few examples of our pre-processed fields (better readability):
\begin{itemize}
    \item politico remember clearly individual admit treason trumpisatraitor mccainsahero johnmccainday
    \item dont know much take compulsive liar trumphours trumpisatraitor
    \item yesterday congress blood donation necessary blah blah blah today congress snub injustice doctor wb shame u call national opposition thuuuuuuu congress westbengal doctorsfightback doctorsprotest soniagandhi rahulgandhi priyankagandhi
    \item opposition shame country gandinaaliabuse
    \item nowthisnews hope one else hire douchebag
    \item oh hell fucktrump
\end{itemize}

Such comments don't fit into the narrative of hate speech, they are critical, cynical or offensive (as the last example, as such comment types regarding Donald John Trump are labelled as hate speech quite frequently). Unfortunately, this data set teaches us that trusting other scientific research without own checking can result in poor quality outputs. This is a serious issue for the scientific community, especially in data science where labelling is done via humans which are prone to errors and which may also add their own opinions/ beliefs and biases during the labelling process. It can't be outlined enough: good quality data labelling is crucial for producing usable and reliable algorithms.

\bibliographystyle{acl_natbib}
\bibliography{nlp}
\appendix
\end{document}